\documentclass[lettersize,journal,twoside]{IEEEtran}
\usepackage{cite}

\ifCLASSINFOpdf
  \usepackage[pdftex]{graphicx}
  \graphicspath{{./figures/}}
  \DeclareGraphicsExtensions{.pdf,.jpg,.jpeg,.png}
\else
\fi
\usepackage{amsmath}
\usepackage{amssymb}
\usepackage{array, multirow}

\ifCLASSOPTIONcompsoc
  \usepackage[caption=false,font=normalsize,labelfont=sf,textfont=sf]{subfig}
\else
  \usepackage[caption=false,font=footnotesize]{subfig}
\fi
\usepackage{url}

\usepackage{hyperref}
\usepackage[nameinlink,capitalize]{cleveref}

\usepackage{eso-pic}

\AddToShipoutPictureFG{%
  \AtPageUpperLeft{%
    \raisebox{-.5\baselineskip}
      {\hspace{16.0\baselineskip}\tiny\sffamily This article has been accepted for publication in IEEE Transactions on Intelligent Transportation Systems. This is the author's version which has not been fully edited and}%
  }%
}
\AddToShipoutPictureFG{%
  \AtPageUpperLeft{%
    \raisebox{-1.2\baselineskip}
      {\hspace{20.9\baselineskip}\tiny\sffamily content may change prior to final publication. Citation information: DOI 10.1109/TITS.2022.3206882}%
  }%
}

\AddToShipoutPictureFG{%
  \AtPageLowerLeft{%
    \raisebox{1.0\baselineskip}
      {\hspace{15.0\baselineskip}\tiny\sffamily \copyright~2022 IEEE. Personal use is permitted, but republication/redistribution requires IEEE permission.	See https://www.ieee.org/publications/rights/index.html for more information.}%
  }%
}

\begin{document}
%
\title{Probabilistic Metamodels for an Efficient Characterization of Complex Driving Scenarios}

%
%
%

\author{Max~Winkelmann,
	Mike~Kohlhoff,
	Hadj~Hamma~Tadjine,~\IEEEmembership{Senior Member,~IEEE,}
	and Steffen~Müller
	\thanks{Manuscript received October 6, 2021; revised April 19, 2022 and September 5, 2022; accepted Month XX, 202X. Date of publication Month XX, 202X. This work was supported in part by IAV GmbH, 10587 Berlin, Germany. The Associate Editor for this article was X. XXXXXX \textit{(Corresponding author: Max Winkelmann.})}
	\thanks{Max Winkelmann and Hadj Hamma Tadjine are with IAV GmbH (e-mail: max.winkelmann@iav.de; hadj.hamma.tadjine@iav.de).}
	\thanks{Mike Kohlhoff was with IAV GmbH. He is now with MHP Management- und IT-Beratung GmbH, 71638 Ludwigsburg, Germany (e-mail: mike.kohlhoff@mhp.com)}
	\thanks{Steffen Müller is with Department of Automotive Engineering, Technische Universität Berlin, 13355 Berlin, Germany (e-mail: steffen.mueller@tu-berlin.de).}
	\thanks{Digital Object Identifier: 10.1109/TITS.XXXX.XXXXXXX}
}

%
%

\markboth{IEEE Transactions on Intelligent Transportation Systems,~Vol.~XX, No.~X, Month~202X}%
{Winkelmann \MakeLowercase{\textit{et al.}}: Probabilistic Metamodels for an Efficient Characterization of Complex Driving Scenarios}
%

\IEEEpubid{\begin{minipage}{\textwidth}\ \\[12pt] \centering
		XXXX--XXXX~\copyright~202X IEEE. Personal use is permitted, but republication/redistribution requires IEEE permission.\\
		See https://www.ieee.org/publications/rights/index.html for more information.
	\end{minipage}}



\maketitle

\begin{abstract}
	To validate the safety of automated vehicles (AV), scenario-based testing aims to systematically describe driving scenarios an AV might encounter. In this process, continuous inputs such as velocities result in an infinite number of possible variations of a scenario. Thus, metamodels are used to perform analyses or to select specific variations for examination. However, despite the safety criticality of AV testing, metamodels are usually seen as a part of an overall approach, and their predictions are not questioned. This paper analyzes the predictive performance of Gaussian processes (GP), deep Gaussian processes, extra-trees, and Bayesian neural networks (BNN), considering four scenarios with 5 to 20 inputs. Building on this, an iterative approach is introduced and evaluated, which allows to efficiently select test cases for common analysis tasks. The results show that regarding predictive performance, the appropriate selection of test cases is more important than the choice of metamodels. However, the choice of metamodels remains crucial: Their great flexibility allows BNNs to benefit from large amounts of data and to model even the most complex scenarios. In contrast, less flexible models like GPs convince with higher reliability. Hence, relevant test cases are best explored using scalable virtual test setups and flexible models. Subsequently, more realistic test setups and more reliable models can be used for targeted testing and validation.
\end{abstract}


%
\IEEEpeerreviewmaketitle

\newcommand{\x}{\boldsymbol{x}}
\newcommand{\y}{\boldsymbol{y}}
\newcommand{\f}{\mathbf{f}}
\newcommand{\X}{\boldsymbol{X}}

\section{Introduction}
\IEEEPARstart{T}{he increasing automation} of road transport promises increased sustainability, more cost-effective mobility, and, above all, greater safety. However, various accidents tragically demonstrated how automated vehicles (AV) can cause fatalities within seconds if they do not sufficiently understand their surroundings~\cite{penmetsa_effects_2021}. Thus, the validation of AVs' safety is in public focus and paramount for their release.

Traditionally, driver assistance systems (SAE level 1-2) are tested using manually defined test cases. This is possible since the systems' operational design domain (ODD) is clearly restricted, and the driver ensures that the system is only used within its ODD or takes control. However, for AV (SAE level 3-5), monitoring by the driver is omitted, and the ODD is less restricted. Hence, it is crucial that tests sufficiently cover the ODD's conditions. A methodologically simple approach to this is to have an AV follow representative routes of human drivers until it is statistically proven that the AV causes fewer fatalities than human drivers. However, this approach not only endangers other road users but requires prohibitive amounts of resources~\cite{kalra_driving_2016}. This is mainly because representative driving is not informative since critical events (e.g., fatalities) are rare.

To enable sufficient coverage and a focus on critical events at the same time, scenario-based testing (SBT)~\cite{riedmaier_survey_2020} aims to systematically describe driving scenarios an AV might encounter. Scenarios are gradually specified: First, a \textit{functional scenario} is described in natural language~\cite{menzel_scenarios_2018}, e.g., 'an AV crosses an intersection'. Then, a corresponding \textit{logical scenario} specifies inputs and their ranges, e.g., the velocity of the AV $v_\mathrm{AV} \in \left[5\ \mathrm{kmh^{-1}}, 50\ \mathrm{kmh^{-1}}\right]$. Finally, \textit{concrete scenarios} with fixed values represent executable test cases, e.g., $v_\mathrm{AV} = 30\ \mathrm{kmh^{-1}}$.

Despite the systematic description of scenarios, safety validation can become intractable. Even with virtual test setups (e.g., 3D simulations), detailed models or perception pipelines can be elaborate, thus limiting the number of concrete scenarios that can be examined. Combined with the rarity of critical events, it is hence necessary to intelligently scan the space of each logical scenario for relevant concrete scenarios~\cite{riedmaier_survey_2020}. For this purpose, techniques from the fields \textit{design of experiments} and \textit{active learning} are applied. Here, a limited number of concrete scenarios is often used to create a metamodel (also called surrogate model) that enables predicting the outcome of further concrete scenarios. Based on predictive analyses, relevant concrete scenarios are then selected for examination.

\IEEEpubidadjcol

Using predictive analyses, metamodels become a crucial part of a safety validation procedure. Non-sufficient performance of a metamodel can lead to critical events being missed or biased risk estimates. Nevertheless, metamodels are usually considered mere components of the process they are embedded in, and their explicit capability to model scenarios has not yet been investigated in detail. To better understand metamodels in the context of SBT, this paper is organized along the following research questions and contributions:

\textit{1)} Which demands does scenario-based testing place on metamodels and which metamodels can satisfy them? This question is answered through a survey of metamodel applications and an analysis of various metamodels' characteristics.

\textit{2)} How do the characteristics of probabilistic metamodels affect their predictive performance and reliability? This question is answered through an experimental benchmark considering four probabilistic metamodels and four driving scenarios with 5 to 20 inputs.

\textit{3)} How does a targeted selection of test cases influence the predictive performance of metamodels? To answer this question, we introduce an iterative approach to efficiently select concrete scenarios and benchmark the resulting metamodels.

\cref{sec:metamodels_in_scenario_based_testing} and \ref{sec:characteristics_of_probabilistic_metamodels} introduce the tasks and characteristics of metamodels. \cref{sec:comparison_metamodels} and \ref{sec:approach} describe the benchmark and iterative approach. Results, conclusions, and future work are discussed in \cref{sec:experiments}, \ref{sec:conclusion}, and \ref{sec:future_work}, respectively.

\section{Metamodels in Scenario-Based Testing}\label{sec:metamodels_in_scenario_based_testing}
\begin{figure}[!t]
	\centering
	\subfloat[The Cut-In Scenario]{\includegraphics[width=1.12in]{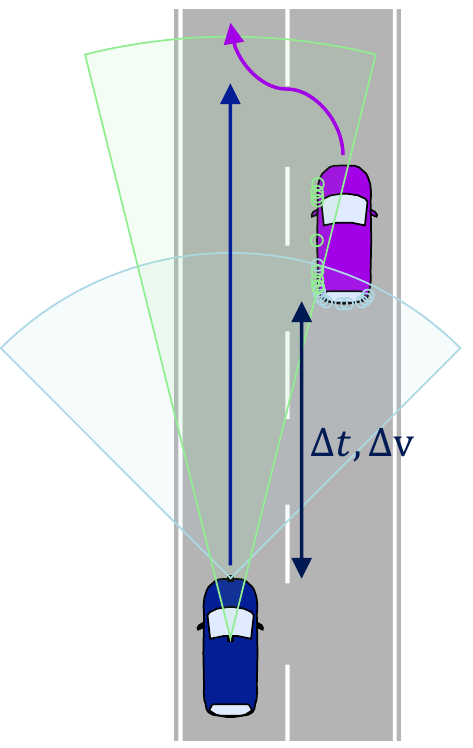}%
		\label{fig:cutin_sketch}}
	\hfil
	\subfloat[Results of the Cut-In Scenario]{\includegraphics[width=2.15in]{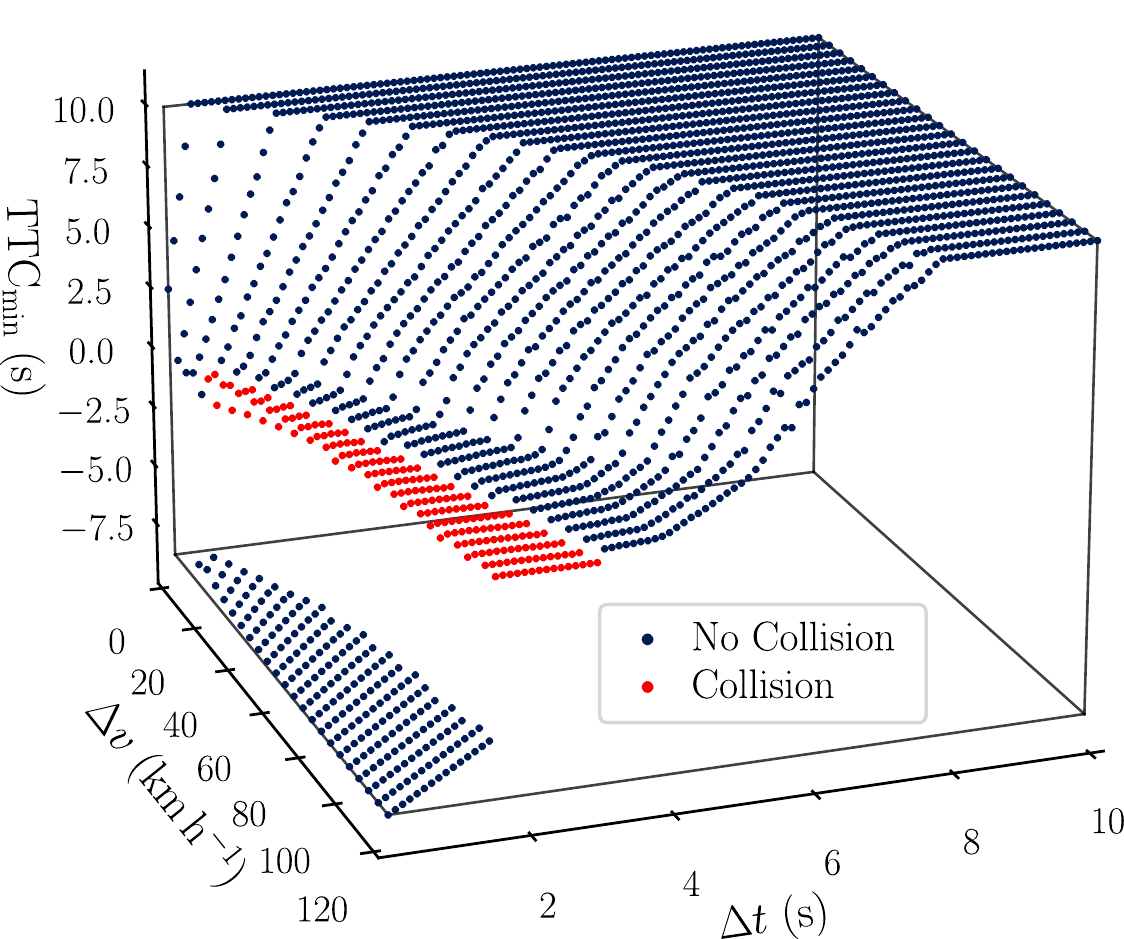}%
		\label{fig:cutin_results}}
	\caption{Depiction of the Cut-In scenario. (a) shows the initial conditions: Starting from a time gap $\Delta t$ and relative velocity $\Delta v$, the vehicle on the right lane cuts into the left lane. The AV on the left lane can brake to avoid collisions. The minimum time to collision $\mathrm{TTC}_\mathrm{min}$ in longitudinal direction is used as the safety metric and shown in (b). Negative values of $\mathrm{TTC}_\mathrm{min}$ indicate that the vehicle on the right cut in behind the AV.
	}
	\label{fig:cutin}
\end{figure}

In SBT, metamodels are typically created for a logical scenario in a test setup $\mathcal{M}$, involving the AV under test and its environment~\cite{kapinski_simulation-based_2016}. The logical scenario defines $N_I$ internal and external influences on the AV's behavior, which we denote by inputs $\x \in \mathbb{R}^{N_I}$. The ranges of the inputs define the input space $I$, containing all possible concrete scenarios. In \cref{fig:cutin}, the inputs are $\Delta t$ as well as $\Delta v$. For each concrete scenario, the AV shows a behavior $\Phi(\mathcal{M}, \boldsymbol{x})$, which is assessed by $y \in \mathbb{R}$ (e.g., the minimum time to collision $\mathrm{TTC}_\mathrm{min}$~\cite{hayward_near-miss_1972}, a common safety metric). Hence, the execution of each concrete scenario gives a sample $(\x, y)$ and $N_S$ samples are a dataset $\X \in \mathbb{R}^{N_S \times N_I}, \y \in \mathbb{R}^{N_S}$ that allows training a metamodel $\widetilde{\mathcal{M}}$.

While various models can act as metamodels and benchmarks are common in fields like computer vision (see, e.g.,~\cite{feng_review_2021}), only few works compare metamodels in SBT~\cite{sun_adaptive_2021, zhang_risk_2022, haq_efficient_2022}; however, they assess their algorithms' overall performance and not the metamodels themselves, which limits generalizability. To gain a deeper understanding, we first analyze the demands common analysis tasks in SBT place on metamodels.

\subsection{The Task of Falsification}\label{subsec:falsification}
Falsification seeks a concrete scenario where safety requirements $\psi$ are violated, i.e., $\Phi(\mathcal{M}, \x) \not\models \psi$~\cite{kapinski_simulation-based_2016}. This often corresponds to minimizing safety metrics below a threshold, as it is done with Gaussian processes (GP) in~\cite{beglerovic_testing_2017} and~\cite{gangopadhyay_identification_2019}. For a similar approach,~\cite{sun_adaptive_2021} compares six metamodels, and extreme gradient boosting performs best. However, in~\cite{zhang_risk_2022} the authors question this result since critical events are frequent in~\cite{sun_adaptive_2021}. In~\cite{beglerovic_testing_2017}, minimization is supported by iteratively focusing the search space around a known minimum. Using local metamodels, which only model subspaces of $I$, can improve the predictive performance since the local behavior tends to be more homogenous and is, therefore, easier to model. Correspondingly, \cite{matinnejad_mil_2014} combines dimensionality reduction, regression trees, and polynomial regression. \cite{haq_efficient_2022} uses an ensemble of a GP, polynomial regression, and a radial basis function network for global search and compares the individual models for local search, observing insignificant differences. \cite{ben_abdessalem_testing_2016} combines a neural network with a genetic algorithm.

For falsification, metamodels need to model a small part of the input space well so that a violation of requirements can be demonstrated with a small number of evaluations of $\mathcal{M}$. Thereby, predictions do not have to be precise but only rank the criticality of potential samples. Nevertheless, metamodels should be able to deal with high $N_I$ to avoid that inputs have to be excluded based on global analyses, even though they may be relevant to provoke critical concrete scenarios. Lastly, a statement regarding metamodel certainty can be essential to avoid falsely concluding that no falsification is possible.

\subsection{The Task of Boundary Identification}\label{subsec:boundary_identification}
A more comprehensive task is the identification of safety boundaries, i.e., the boundaries between the fulfillment $\Phi(\mathcal{M}, \x) \models \psi$ and violation $\Phi(\mathcal{M}, \x) \not\models \psi$ of requirements. Here,~\cite{zhou_safety_2018} and~\cite{batsch_performance_2019} use GP classification models and reason that high classification uncertainty indicates the safety boundaries. Since classification models require observations of the fulfillment and violation of requirements, \cite{batsch_scenario_2021} and \cite{sun_adaptive_2021} use GP regression models to select samples resulting in a specified $y$.

For boundary identification, metamodels must perform well across a larger part of the input space than for falsification. To place samples on the safety boundary, predictions have to be precise. Additionally, predicted uncertainties are valuable to place samples where their classification is ambiguous.

\subsection{The Task of Estimation}
Estimation aims to estimate the probability of requirement violations given the likelihood of concrete scenarios, i.e., $P(\Phi(\mathcal{M}, \x) \not\models \psi)\ |\ \mathcal{M},\ p(\x),\ \psi)$. In~\cite{huang_towards_2017} and~\cite{winkelmann_transfer_2022}, GPs are used to assess potential samples within an importance sampling approach. For a similar approach,~\cite{zhang_risk_2022} compares six metamodels, and an inverse distance weighted model performs best. \cite{feng_testing_2022} combines GP classification and regression to model dissimilarities between an AV and its model. \cite{sinha_neural_2020} uses a GP to obtain gradients for the behavior of an examined system.

For estimation, metamodels must properly model multiple clusters of critical samples. Furthermore, it can be essential that the models' uncertainties are well-calibrated, i.e., they not only rank the expected criticality of samples, but the expected criticality matches the actual frequency of critical events.

\section{Characteristics of Probabilistic Metamodels}\label{sec:characteristics_of_probabilistic_metamodels}
The varying results in~\cite{sun_adaptive_2021, zhang_risk_2022, haq_efficient_2022} do not allow generalizable insights regarding metamodels. However, there are manifold demands on metamodels: Large input spaces have to be dealt with, predicted uncertainties are valuable to guide exploration, and particularly for estimation and termination criteria, model calibration is crucial. Hence, probabilistic models such as GPs are suitable models since they explicitly consider uncertainties caused by limited training data and the stochastic outcomes of scenarios. In fact, most of the approaches discussed in \cref{sec:metamodels_in_scenario_based_testing} use GPs. However, most examined scenarios are simple with regard to the number of inputs ($N_I \leq 3$), \cite{sinha_neural_2020} being an exception with $N_I = 24$. Therefore, it is unclear how well GP scale with respect to $N_I$~\cite{corso_survey_2022}. Classic alternatives to GPs are random forests (RF) and extra-trees (ET). Furthermore, recent approaches like Bayesian neural networks (BNN) and deep Gaussian processes (DGP) achieved promising results for many tasks~\cite{salimbeni_doubly_2017} but have not been applied in SBT.

In the following, we discuss the properties of the mentioned metamodels that are relevant for use in SBT. Since our benchmark only investigates a limited number of logical scenarios, the theoretical analysis helps to generalize experimental insights. Generally, probabilistic metamodels need to consider two types of uncertainties. \textit{Aleatoric} uncertainty (from Latin 'alea', dice) accounts for noise in the training data, e.g., caused by sensor noise, and does not reduce with more data~\cite{der_kiureghian_aleatory_2009}. \textit{Epistemic} uncertainty (from Greek 'episteme', knowledge) accounts for the uncertainty about a model's parameters and reduces with more data~\cite{der_kiureghian_aleatory_2009}.

\subsection{Gaussian Processes}\label{subsec:GP}
The usage of GPs~\cite{rasmussen_gaussian_2005} (also called Kriging) is often motivated by their clear mathematical definition. GPs' non-parametric nature scales the model complexity to the amount of data, and choosing hyperparameters is relatively simple. Furthermore, kernel functions allow the introduction of prior knowledge~\cite{rasmussen_gaussian_2005}. However, the definition of kernel functions suitable for SBT is an open problem~\cite{feng_testing_2022}. All approaches in \cref{sec:metamodels_in_scenario_based_testing} use stationary kernels, for which the epistemic uncertainty of a predicted sample is determined primarily by its distance to observed samples. This assumption can lead to erroneous conclusions: E.g., the $\mathrm{TTC}_\mathrm{min}$ of a cut-in at $\Delta t = 10\ \mathrm{s}$ and $8\ \mathrm{s}$ might not differ much, but it can differ greatly at $\Delta t = 3\ \mathrm{s}$ and $1\ \mathrm{s}$ (see \cref{fig:cutin}). Furthermore, GPs typically consider \textit{homoscedastic} aleatoric uncertainty, i.e., it is assumed that all concrete scenarios have equal variance. However, concrete scenarios along an AV's decision boundaries might be more stochastic than other concrete scenarios.

Thus, for GPs, ease of use contrasts with limited capabilities in modeling non-linear, inhomogeneous system behaviors.

\subsection{Random Forests and Extra-Trees}\label{subsec:ET}
Rather than trusting the prediction of a single complex model, the motivation behind ensemble models is to have many simple models vote on a prediction. Here, RFs~\cite{breiman_random_2001} and ETs~\cite{geurts_extremely_2006} are popular approaches since the individual decision trees can be trained within short time and allow understanding the influences of different inputs~\cite{matinnejad_mil_2014}. However, decision trees are not well-suited to model interactions between inputs.

Thus, transparency and speed face the constraint that system behaviors are explained mainly by individual inputs rather than their interaction. We investigate ETs since they were superior to RFs during hyperparameter tuning (see \cref{sec:experiments}).

\subsection{Bayesian Neural Networks}\label{subsec:BNN}
Popularized by the rise of neural networks (NN), the idea behind BNNs~\cite{neal_bayesian_1996} is to consider all parameters not as deterministic values but as probability distributions. Substituting discrete outputs $\hat{y}$ by a mean $\mu$ and variance $\sigma^2$, the aleatoric uncertainty is \textit{heteroscedastic}, i.e., it depends on $\x$ and can be explicitly learned by the BNN. To consider epistemic uncertainty, ensemble models can be used~\cite{lakshminarayanan_simple_2017}. However, their training is very costly. Hence, we only train one BNN and use Monte Carlo (MC) dropout, where multiple predictions are generated with different neurons being deactivated~\cite{gal_concrete_2017}.

Due to their high flexibility, BNNs can represent non-linear systems well. However, BNNs can be sensitive to hyperparameters, and their predictions are hardly traceable.

\subsection{Deep Gaussian Processes}\label{subsec:DGP}
Combining the concepts of GPs and NNs, DGPs are a network of GPs~\cite{salimbeni_doubly_2017}, allowing for hierarchical learning. Since GPs themselves are capable models, DGPs only contain a few layers and GPs per layer. Furthermore, DGPs are based on variational GPs, allowing for training with vast quantities of data without the large memory requirements of traditional GPs.

Hence, DGPs can capture more complex relationships than conventional GPs while remaining relatively transparent.

\section{Experimental Benchmark of Metamodels}\label{sec:comparison_metamodels}

\begin{figure*}[!t]
	\centering
	\includegraphics[width=7in]{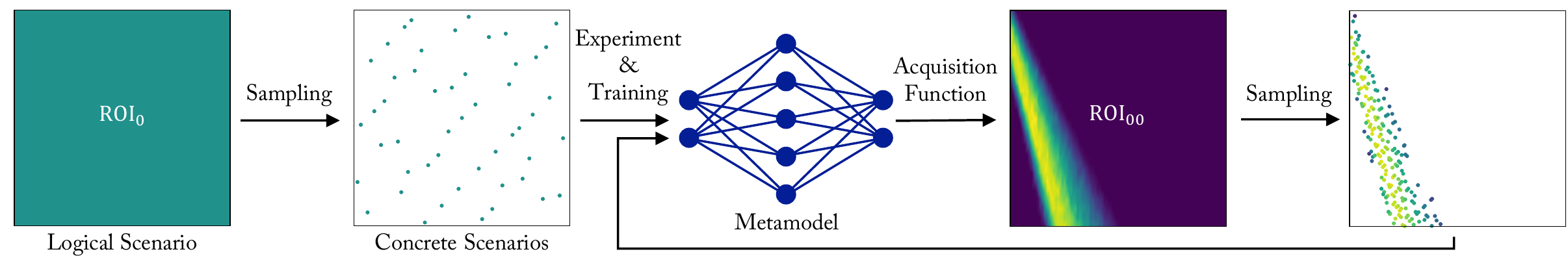}
	\caption{Illustration of our iterative approach. A logical scenario spans a region of interest (ROI) over its respective inputs. From the ROI, concrete scenarios are sampled, which are initially evenly distributed. These scenarios are executed and used to train a metamodel. To generate new samples in a targeted manner, the metamodel is combined with an acquisition function. This results in a new ROI, which is a distribution of concrete scenarios over the input space. Sampling from this distribution, the density of concrete scenarios corresponds to the value of the acquisition function. The concrete scenarios are then carried out, the results are added to the dataset, and a refined metamodel is trained. This procedure can be repeated to iteratively refine the metamodel.}
	\label{fig:approach}
\end{figure*}

To investigate how the metamodels' characteristics affect their predictive performance, we examine them in a benchmark. Ultimately, the goal of the benchmark is to identify a metamodel suited for a wide range of scenario-based tests. For the purpose of generalization, we, therefore, consider five logical scenarios that differ in complexity and type, i.e., number of inputs, driving tasks, and safety metrics.

\subsection{Examined Scenarios}
To balance data expense and realism, the scenarios are modeled and simulated in a 2D environment (IAV Scene Suite), and sensor signals are overlaid with Gaussian noise. This enabled us to create tabular datasets with 10 Million concrete scenarios that can be used for future studies. In the following, we provide short descriptions of the scenarios. More details and references are given in the \hyperref[sec:dataset]{Appendix}.

The longitudinal scenarios \textit{Emergency Braking} ($N_I = 5$) and \textit{Following} ($N_I = 13$) evaluate a combination of adaptive cruise control and automatic emergency braking. Here, a $\mathrm{TTC_{min}} < 0.1\ \mathrm{s}$ is considered critical. The scenario \textit{Lane Keeping} ($N_I = 15$) evaluates a lane keeping controller based on the maximum absolute deviation from a desired trajectory; $\mathrm{p_{err\ abs\ max}} > 3\ \mathrm{m}$ is considered critical. The scenario \textit{Parking} ($N_I = 20$) models a typical parking lot and uses a safety metric similar to the one in~\cite{beglerovic_testing_2017}: Positive values indicate the minimum distance, $0$ is used for collisions after the AV stopped, and negative values indicate the velocity of the AV in the moment of a collision. Here, crashes at more than $10\ \mathrm{kmh^{-1}}$ are considered critical, i.e., $\mathrm{Criterion} < -10$. The dataset also includes the scenario \textit{Cut-In} shown in \cref{fig:cutin}.

\subsection{Experimental Setup}\label{subsec:experimental_setup_metamodel}
In all experiments, a metamodel $\widetilde{\mathcal{M}}$ for a logical scenario is built based on $N_S$ concrete scenarios. Without prior knowledge or assumptions about the behavior, the whole input space $I$ of the logical scenario is equally important, and we call it region of interest (ROI) (see \cref{fig:approach}). From the concrete scenarios within the ROI, a representative selection can be sampled using the Sobol sequence~\cite{sobol_distribution_1967} which ensures reproducibility and samples of consistent quality, also in higher dimensions~\cite{morokoff_quasi-monte_1995}. The generated concrete scenarios $\X$ are then carried out to obtain the results $\y$, and $\widetilde{\mathcal{M}}$ is trained.

\subsection{Evaluation Criteria}\label{subsec:evaluation_criteria}
To assess the metamodel $\widetilde{\mathcal{M}}$, a reference dataset is utilized, which, to avoid correlations between the training and reference data, consists of pseudo-randomly distributed samples. The true values $y$ are compared to the predicted $\mu(\x)$ and $\sigma(\x)$. Since we consider probabilistic metamodels, our primary quality indicator is the likelihood in \cref{eq:likelihood}.
\begin{equation} \label{eq:likelihood}
	p(\y \mid \X, \widetilde{\mathcal{M}}) = \sqrt[N_S]{\prod_{i=1}^{N_S} \frac{1}{\sigma(\x_i) \sqrt {2\pi } }e^{{ - \left( {y_i - \mu(\x_i) } \right)^2 } \mathord{\left/ {\vphantom {{ - \left( {y_i - \f(\x_i) } \right)^2 } {2\sigma(\x_i) ^2 }}} \right. \kern-\nulldelimiterspace} {2\sigma(\x_i) ^2 }}}
\end{equation}
For the likelihood, deviations between $\mu(\x)$ and $y$ are acceptable if they come with a suitable $\sigma(\x)$. The likelihood is calculated in log-space to increase numeric stability.

A secondary quality indicator is the root mean squared error (RMSE) in \cref{eq:rmse}, which does not take into account $\sigma(\x)$.
\begin{equation} \label{eq:rmse}
	\mathrm{RMSE}(\y \mid \X, \widetilde{\mathcal{M}}) = \sqrt{\frac{1}{N_S} \sum_{i=1}^{N_S} (y_i - \mu(\x_i))^2}
\end{equation}

To assess efficiency, we compare how many concrete scenarios are necessary to achieve a certain likelihood or RMSE. To assess the predictions in more depth, we utilize scatter plots, precision-recall curves, and reliability curves.

\section{Targeted Selection of Samples}\label{sec:approach}
While evenly distributed samples are suitable for the initial examination of unknown systems, the tasks discussed in \cref{sec:metamodels_in_scenario_based_testing} require efficient learning strategies. In part, the discussed approaches use techniques that do not scale well to scenarios modeled with many inputs, e.g., choosing samples from a full-factorial set of possible samples. In the following, we demonstrate how samples can be selected in a way that scales to scenarios with high $N_I$. Our approach allows executing all tasks described in \cref{sec:metamodels_in_scenario_based_testing}. Hence, it can serve as a basis for a scalable testing framework and as a baseline for future work using our dataset (see \hyperref[sec:dataset]{Appendix}).

As illustrated in \cref{fig:approach}, a trained metamodel can be combined with an acquisition function $a(\widetilde{\mathcal{M}}, \x)$ resulting in an updated ROI. Sampling from the updated ROI, the density of samples will correspond to the values of $a(\widetilde{\mathcal{M}}, \x)$. In the following, we discuss how the definition of $a(\widetilde{\mathcal{M}}, \x)$ allows performing the tasks discussed in \cref{sec:metamodels_in_scenario_based_testing}. Since falsification is not necessarily informative~\cite{shalev-shwartz_formal_2017} and for estimation, the distribution of concrete scenarios $p(\x)$ has to be known, we evaluate our approach for the task of boundary identification.

\begin{figure*}[!t]
	\centering
	\includegraphics[width=7in]{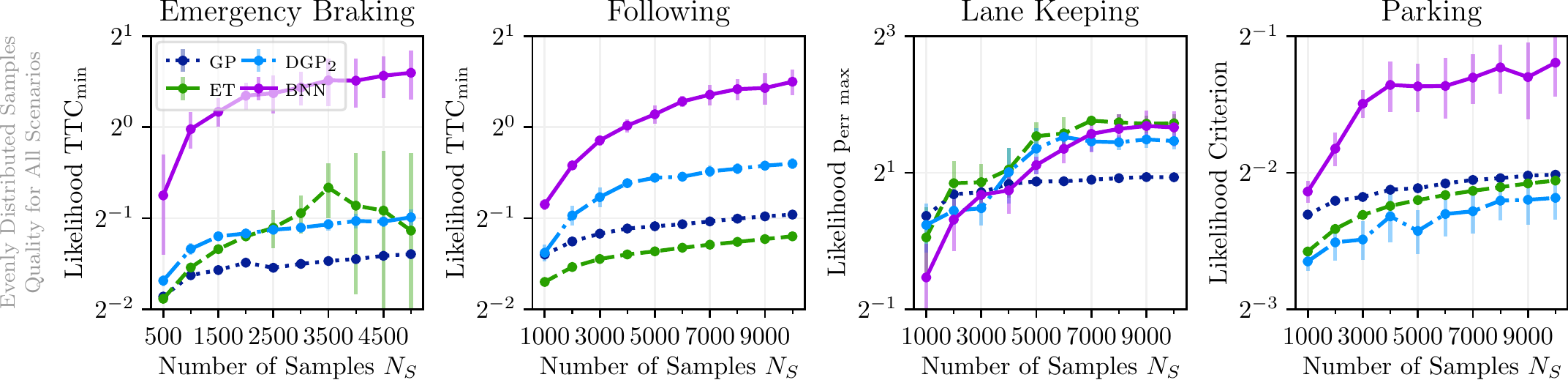}
	\caption{Achieved likelihoods for all test cases using evenly distributed samples. The likelihoods assess how well the metamodels can predict the concrete scenarios' safety metrics of a reference set (see \autoref{subsec:evaluation_criteria}). Dots represent the mean and bars the standard deviation over 50 repetitions. Generally, an increasing number of samples $N_S$ leads to higher (better) likelihoods. While for small $N_S$, the differences between the models can be marginal, some models show a logarithmic relationship between $N_S$ and their likelihood, indicating that they would require orders of magnitude more data to achieve the same likelihood as superior models. The BNN benefits from high $N_S$ and consistently reaches the highest likelihoods for 3 out of 4 logical scenarios.}
	\label{fig:evenly_distributed_likelihood_all}
\end{figure*}

\begin{figure*}[!t]
	\centering
	\includegraphics[width=7in]{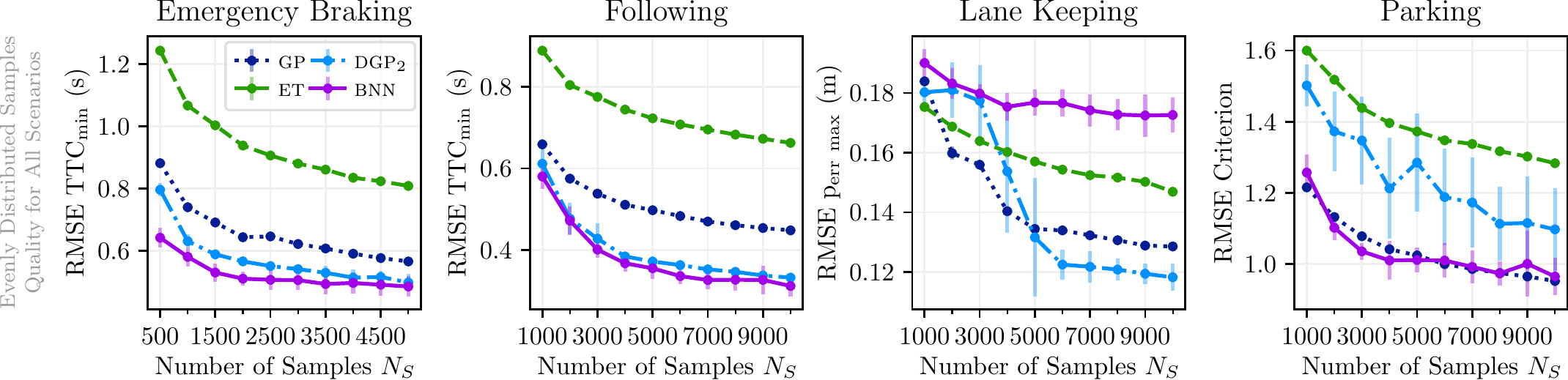}
	\caption{Achieved RMSEs (root mean squared errors) for all test cases using evenly distributed samples. The RMSEs assess how well the metamodels can predict the concrete scenarios' safety metrics of a reference set, considering only the predicted mean and not the models' predicted uncertainties (see \autoref{subsec:evaluation_criteria}). Dots represent the mean and bars the standard deviation over 50 repetitions. While the RMSEs reduce (improve) with increasing $N_S$, the logarithmic slopes suggest that some models require orders of magnitude more data to reach the RMSEs of others.}
	\label{fig:evenly_distributed_rmse_all}
\end{figure*}

\subsection{Boundary Identification}\label{subsec:characterize_borders}
A violation of requirements $\psi$ can often be characterized as an output $y$ lying within a critical interval $M_{\mathrm{c}}$. For example, in \cref{fig:cutin} we may define that a $\mathrm{TTC}_\mathrm{min} \in \left[-5\ \mathrm{s}, 0.5\ \mathrm{s}\right]$ corresponds to a violation. Probabilistic predictions allow calculating the probability that $y \in M_{\mathrm{c}}$ using the cumulative distribution function. As discussed in \cref{subsec:boundary_identification}, examining samples for which the classification is ambiguous should reveal the safety boundaries. Such samples can be generated using the acquisition function $a_{\mathrm{BI}}$ in \cref{eq:a_bi}, which is calculated in log-space to ensure numerical stability.
\begin{equation} \label{eq:a_bi}
	a_{\mathrm{BI}}(\x, \widetilde{\mathcal{M}}, M_{\mathrm{c}}) = 1 - 2\left| P(y \in M_{\mathrm{c}} \mid \mu(\x), \sigma(\x))-0.5 \right|
\end{equation}
Since the resulting ROI can be arbitrarily complex, new samples are drawn using rejection sampling~\cite{von_neumann_various_1951}. Here, using the same Sobol sequence across iterations ensures that new samples are placed between existing ones. The number of samples is freely selectable, allowing to balance the expenses of carrying out test cases and metamodel training. The generated samples are then executed, added to the dataset, and a new metamodel is trained. Here, the first 20 \% of the samples are used as validation and the last 80 \% as training data to preserve the distribution of the Sobol sequence. Even without observing critical behavior, the continuous output of the regression models allows to progressively explore and model the safety boundaries over multiple iterations.

\subsection{Falsification and Estimation}\label{subsec:characterize_interesting}
Falsification can be realized using acquisition functions from Bayesian optimization~\cite{beglerovic_testing_2017, gangopadhyay_identification_2019}. For estimation, metamodel-based importance sampling~\cite{dubourg_metamodel-based_2013} can be applied, as shown in \cite{winkelmann_transfer_2022}. Here, the acquisition function $a_{\mathrm{IS}}$ considers the distribution of concrete scenarios $p(\x)$ as shown in \cref{eq:a_is}.
\begin{equation} \label{eq:a_is}
	a_{\mathrm{IS}}(\x, \widetilde{\mathcal{M}}, M_{\mathrm{c}}, p) = P(y \in M_{\mathrm{c}} \mid \mu(\x), \sigma(\x))\ p(\x)
\end{equation}

\section{Experimental Results}\label{sec:experiments}
To investigate the metamodels' performance as well as the influence of the acquisition function, we apply our approach for different tuples of logical scenarios, metamodels, and acquisition functions. For each tuple, the number of iterations of our iterative approach and the number of samples per iteration, have to be chosen. We use a fixed number of 10 iterations and, according to the complexity of the scenarios, 50 samples per iteration for the scenario Cut-In, 500 for the scenario Emergency Braking, and 1000 for the remaining scenarios. To investigate the variance of the results, each tuple was executed 50 times with different random seeds.

To fully exploit the models' potentials and at the same time investigate models that generalize well to future applications, we used Bayesian optimization to obtain one set of hyperparameters per model (see \cref{tab:hyperparameters}). For a given hyperparameter set, the corresponding model had to model all five logical scenarios using evenly distributed samples for the initial and final $N_S$. Then, the likelihoods of the reference datasets were multiplied, and the product was maximized. Due to its low $N_I$, we omit the scenario Cut-In in the following examinations.

\begin{figure*}[!t]
	\centering
	\includegraphics[width=6.2in]{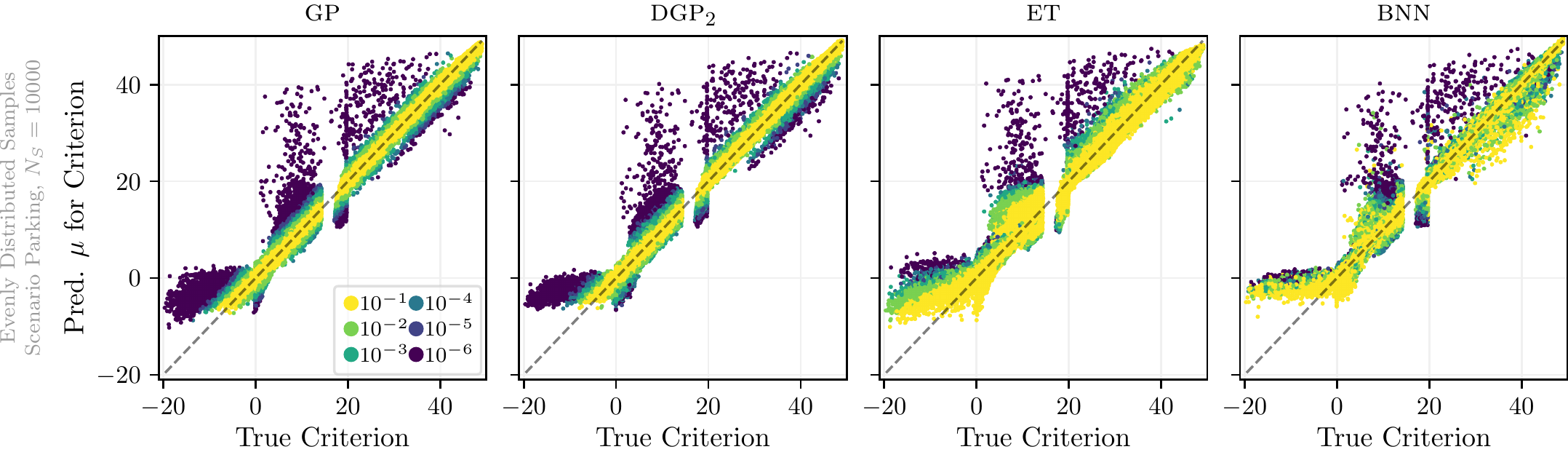}
	\caption{Predictions for the scenario Parking based on 10 iterations of evenly distributed samples resulting in $N_S = 10000$. The $x$-axis shows the true safety metrics of a reference dataset; the $y$-axis shows the metamodels' predicted means. The predictions are colored based on the residual risk that needs to be aimed for to cover the true outcome given the model, e.g., if a residual risk of $10^{-2}$ is aimed for, an interval of $\pm 2.58 \sigma$ around $\mu$ needs to be considered and all true outcomes lying within this interval are colored accordingly. The figure shows the predictions of the run that achieved the likelihood closest to the median likelihood of the 50 repetitions. All models capture the general behavior but give many unlikely predictions.}
	\label{fig:scatter_equidistributed}
\end{figure*}

\begin{figure*}[!t]
	\centering
	\includegraphics[width=6.2in]{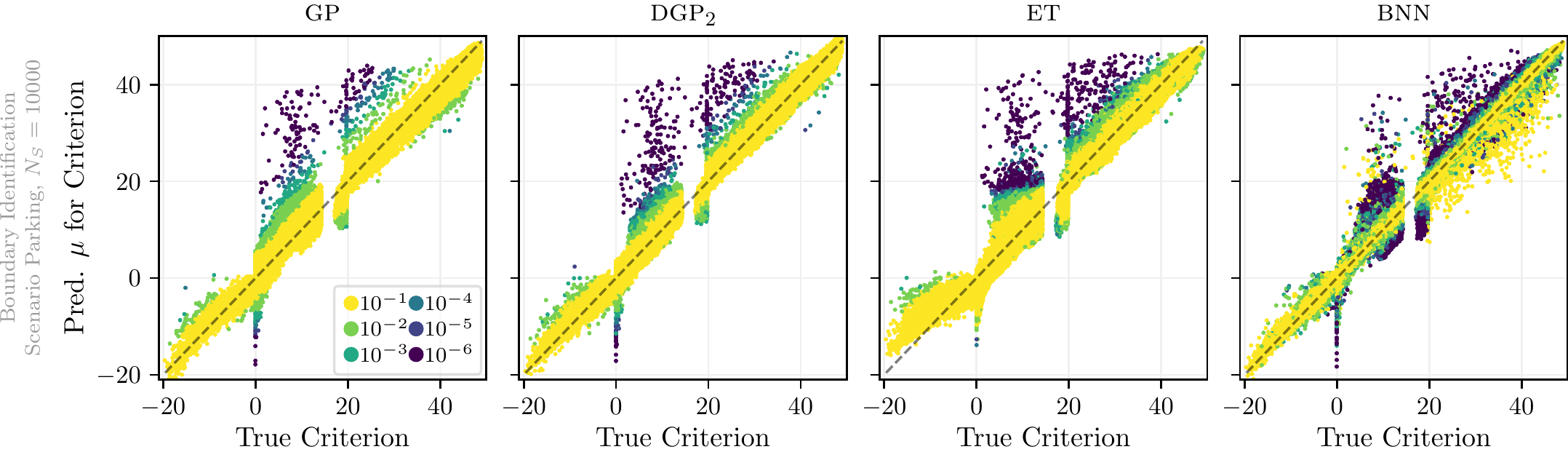}
	\caption{Predictions for the scenario Parking based on 10 iterations using the acquisition function for boundary identification $a_{\mathrm{BI}}$ resulting in $N_S = 10000$. The colors and test run are chosen the same way as in the figure above. Compared to the selection of evenly distributed samples shown above, it can be seen that the critical samples with $\mathrm{Criterion} < -10$ are predicted more accurately, and the fraction of unlikely predictions is significantly reduced.}
	\label{fig:scatter_characterize_borders}
\end{figure*}

\subsection{Global Model Quality Using Evenly Distributed Samples}
First, we examine the metamodels without a targeted selection of samples and evaluate the global quality, i.e., the predictive performance for all test cases. We generate evenly distributed samples using a constant acquisition value $a_{\mathrm{ES}} = 1$. The achieved likelihoods are shown in \cref{fig:evenly_distributed_likelihood_all}. For 3 out of 4 logical scenarios, the BNN reaches the highest likelihoods for all $N_S$. Interestingly, the RMSEs in \cref{fig:evenly_distributed_rmse_all} give a more balanced impression. Despite lower likelihoods, the GP and DGP can often achieve the same or lower RMSEs than the BNN and ET. As discussed in \cref{subsec:evaluation_criteria}, this indicates that the errors of the ET's and BNN's predictions are accompanied by suitable uncertainties. This can be supported considering \cref{fig:scatter_equidistributed}, especially for the samples where crashes occurred, i.e., $\mathrm{Criterion} \leq 0$. For the GP and DGP, the predictions' likelihoods are highly correlated to their absolute error (the vertical distance to the diagonal). According to \cref{eq:likelihood}, this indicates that the predicted uncertainties are homogeneous, which, based on the discussion in \cref{subsec:GP} can be attributed to how GPs model uncertainty. In contrast, the ET and BNN potentially predict higher uncertainties for samples with higher errors, i.e., their uncertainties are heterogeneous.

The potential drawbacks of heterogeneous uncertainties are evident for the BNN in the scenario Lane Keeping. For specific concrete scenarios, the lane keeping controller gets unstable, resulting in high $\mathrm{p_{err\ abs\ max}}$. The BNN is flexible enough to interpret these rare events as noise. This way, the large predicted $\sigma(\x)$ result in competitive likelihoods. However, the predicted $\mu(\x)$ do not follow $y$, resulting in poor RMSEs.

In terms of global likelihoods and RMSEs, the BNN achieves the best performance overall. However, the high fraction of unlikely predictions in \cref{fig:scatter_equidistributed} indicates that none of the models is well-calibrated, which is discussed in \cref{subsec:results_characterize_boundaries}. It is also striking that critical samples are predicted poorly by all models. However, considering that critical samples are rare (for \cref{fig:scatter_equidistributed}, the models observe about 10), the poor predictions are plausible and confirm that for the tasks discussed in \cref{sec:metamodels_in_scenario_based_testing}, samples must be selected efficiently.

\begin{figure*}[!t]
	\centering
	\includegraphics[width=7in]{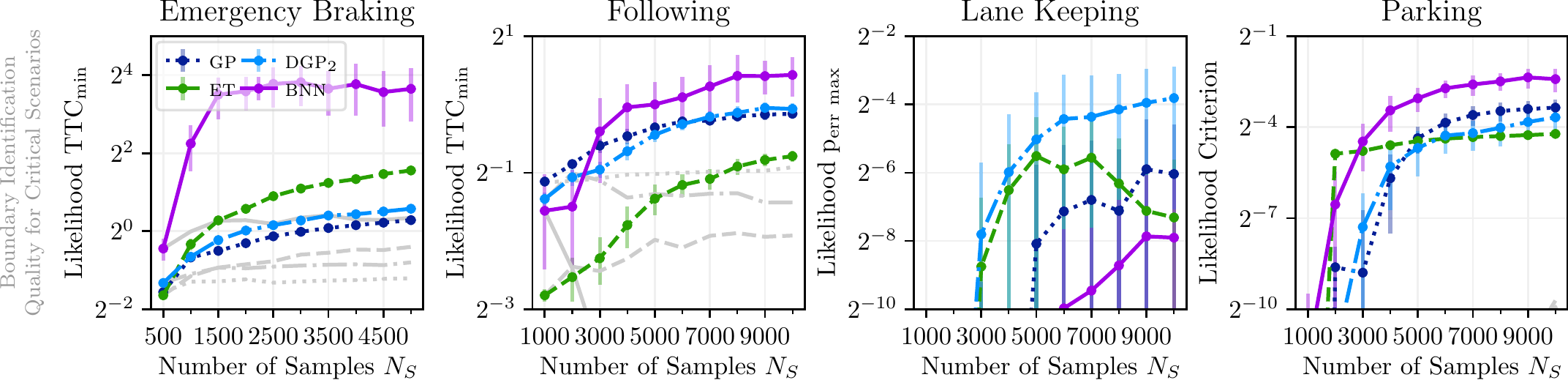}
	\caption{Achieved likelihoods for the critical test cases applying the acquisition function for boundary identification $a_{\mathrm{BI}}$. Dots represent the mean and the bars the standard deviation over 50 repetitions. The likelihoods for evenly distributed samples are shown in gray. For the scenarios Lane Keeping and Parking, these remain below the depicted values. The application of the acquisition function significantly improves the likelihoods of the critical samples.}
	\label{fig:characterize_borders_likelihood_interesting_only}
\end{figure*}

\begin{figure*}[!t]
	\centering
	\includegraphics[width=7in]{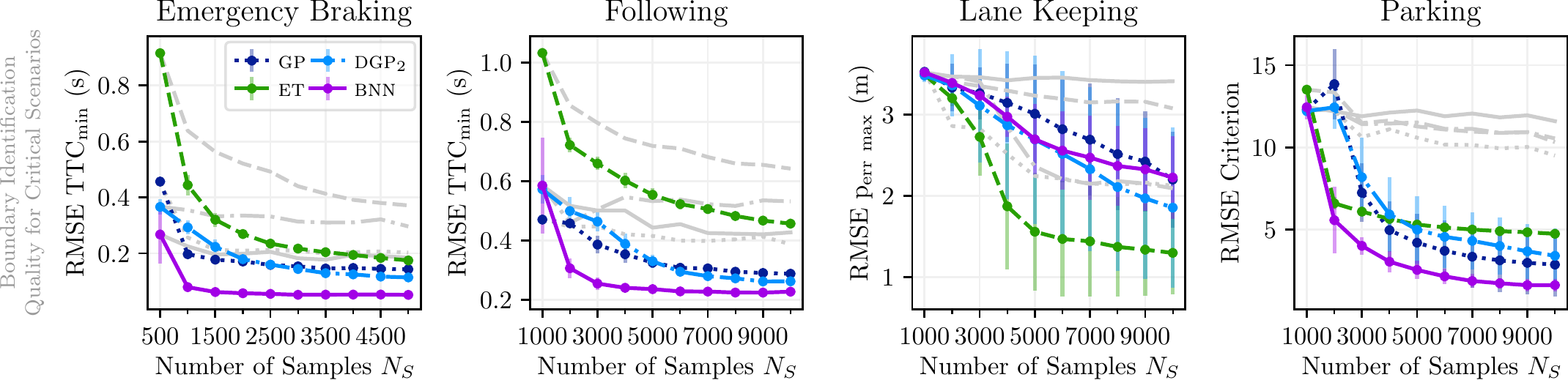}
	\caption{Achieved RMSEs for the critical test cases applying the acquisition function for boundary identification $a_{\mathrm{BI}}$. Dots represent the mean and the bars the standard deviation over 50 repetitions. The RMSEs for evenly distributed samples are shown in gray. Applying the acquisition function, the RMSEs can be significantly reduced. In some cases, it converges after a few iterations indicating that further improvement is difficult and a small $N_S$ can be sufficient.}
	\label{fig:characterize_borders_rmse_interesting_only}
\end{figure*}

\subsection{Model Quality for Critical Concrete Scenarios}
We now investigate how the acquisition function for boundary identification $a_{\mathrm{BI}}$ (see \cref{eq:a_bi}) influences the predictions of critical samples; \cref{fig:scatter_characterize_borders} gives a visual impression. In all following figures, the results for evenly (quasi-randomly) distributed samples are shown in gray and serve as a baseline.

Since precise predictions of the critical samples are essential for all tasks discussed in \cref{sec:metamodels_in_scenario_based_testing}, we evaluate the achieved likelihoods in \cref{fig:characterize_borders_likelihood_interesting_only}. Applying the acquisition function can greatly increase the likelihood of critical samples. This is also the case for the RMSEs shown in \cref{fig:characterize_borders_rmse_interesting_only}.

Overall, it shows that often the best metamodel trained with evenly distributed samples cannot keep up with the worst one using the acquisition function. Thus, the influence of the acquisition function appears to be greater than that of the metamodel. Still, the BNN achieves the highest likelihoods and lowest RMSEs for 3 out of the 4 logical scenarios.

\subsection{Classification Performance}\label{subsec:results_characterize_boundaries}
\begin{figure*}[!t]
	\centering
	\includegraphics[width=6in]{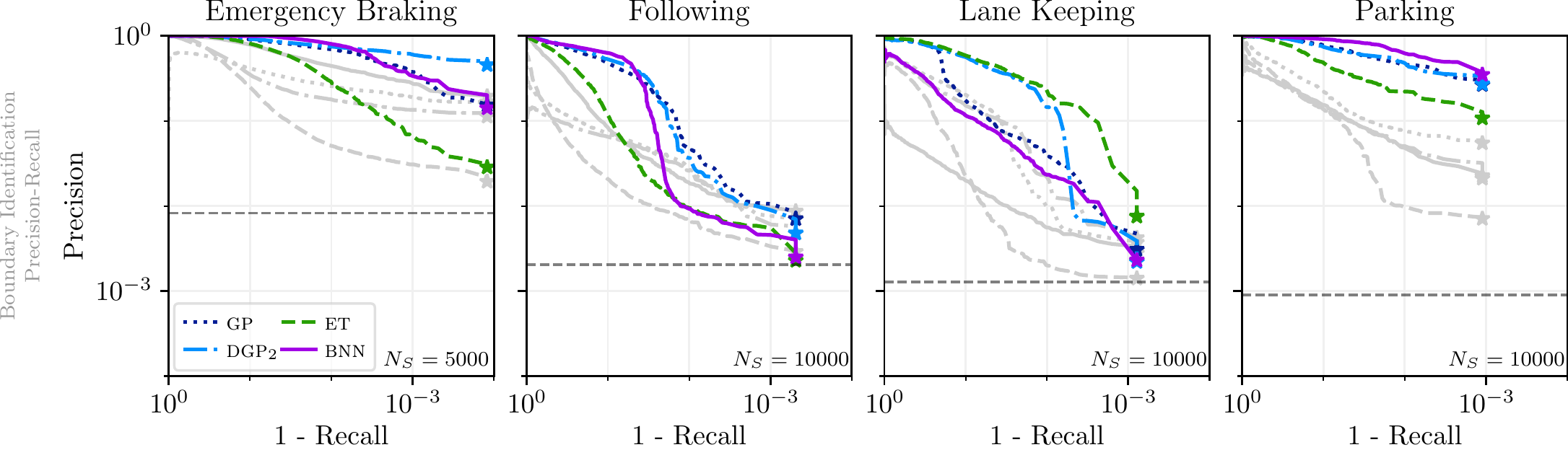}
	\caption{Precision-recall curves applying the acquisition function for boundary identification. The horizontal gray lines indicate the frequency of critical samples for evenly distributed samples. The stars indicate the point where all critical samples are found. The results for evenly distributed samples are shown in gray. The figure shows the data of the run that achieved the likelihood closest to the median likelihood of the 50 repetitions. Applying the acquisition function, especially the precision for lower recalls can be improved. To identify all critical samples, the precision can only be improved in some cases.}
	\label{fig:precision-recall_characterize_borders}
\end{figure*}

For boundary identification (see \cref{subsec:boundary_identification}), it is crucial that the metamodels can distinguish critical and non-critical samples. To conduct such binary classification, a threshold for $P(y \in M_{\mathrm{c}} \mid \mu(\x), \sigma(\x))$ can be defined, above which a sample is considered critical; a high threshold reduces false positives, while a low threshold helps to identify more critical samples. The precision-recall curves in \cref{fig:precision-recall_characterize_borders} show that the application of the acquisition function $a_{\mathrm{BI}}$ mostly increases the precision and therefore improves the classification performance. Considering a recall of 90 \%, the GP, DGP, and BNN achieve precisions that increase the frequency of critical samples by two orders of magnitudes or more for all logical scenarios. However, to identify the remaining 10 \% of critical samples as such, the precision sometimes converges to the natural frequency of critical samples. In rare cases, the application of the acquisition function reduces the precision.

The results above confirm that our approach is suitable to characterize safety boundaries. However, while identifying \textit{most} of the safety boundaries is done very efficiently, a \textit{complete} identification can not always be achieved.

\begin{figure*}[!t]
	\centering
	\includegraphics[width=6in]{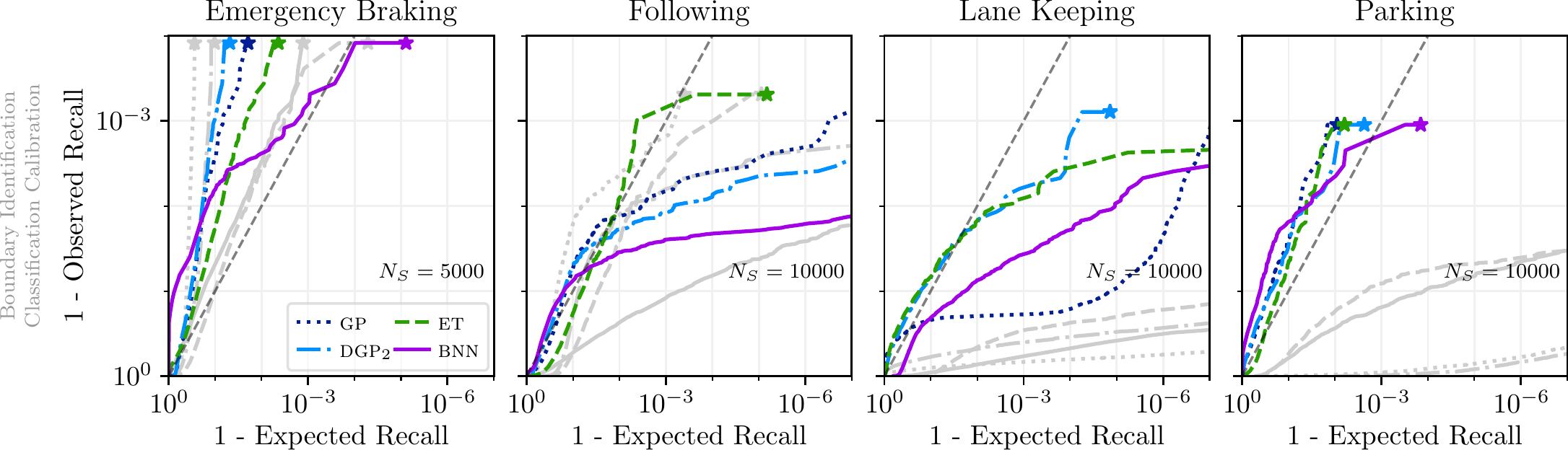}
	\caption{Reliability curves applying the acquisition function for boundary identification. Data and annotations are chosen as in the figure above. The gray diagonal indicates perfect calibration. In many cases (especially for the scenarios Lane Keeping and Parking), the application of the acquisition function can improve the calibration. Nevertheless, the metamodels are often underconfident for lower expected recalls and overconfident for higher expected recalls. This indicates that a small fraction of the predictions can be very unlikely and cause extreme differences between the expected and observed recall.}
	\label{fig:calibration_all_classification}
\end{figure*}

\subsection{Model Calibration}
Examining scenarios with aleatoric uncertainty and limited data, a complete characterization can not be expected. It is, therefore, essential to know the degree to which a scenario is characterized. To this end, the classification calibration is particularly relevant in practice and examined in \cref{fig:calibration_all_classification}. The acquisition function can mostly improve the calibration. The ET did not perform well with previous performance indicators but achieves good calibrations. In contrast, the BNN's calibration is often worse than that of the other models. Here, it becomes visible that due to the varying degrees of flexibility and the associated bias-variance tradeoff, different models shine with respect to different performance indicators.

\subsection{Running Times of Model Training and Sampling}
For all models, the training duration $t_\mathrm{train}$ depends mainly on $N_S$. Hence, in \cref{tab:hyperparameters} we list $t_\mathrm{train}$ for $N_S = 500$ (averaged over the scenarios Cut-In and Emergency Brake) and $N_S = 10^4$ (averaged over the remaining scenarios). The ET is the fastest model (without a GPU). All other models use an NVIDIA GeForce RTX 2080 Ti. Here, the GP is slower than the ET but still much faster than the BNN and DGP.

The duration of the sampling process for the boundary identification task $t_\mathrm{sampling}$ is stated per sample. $t_\mathrm{sampling}$ depends mainly on the frequency of critical samples. If critical samples are rare, more potential samples have to be generated until suitable samples are identified. Hence, $t_\mathrm{sampling}$ is stated for the scenarios Lane Keeping and Parking (where critical samples are rare) and averaged over all iterations. For sampling, the differences are smaller than for training. Sampling with the ET is the fastest, followed by the BNN, GP, and DGP.

To conclude, the more time-consuming or expensive individual test cases are, the more likely the effort of model training and sampling will be negligible compared to the effort of additional test cases. For applications with small batches (e.g., for iterative proving ground tests), fast models like ET or GP may have to be used since long running times are prohibitive.

\section{Conclusion}\label{sec:conclusion}
This paper provided an in-depth analysis of metamodels for the characterization of driving scenarios. Regarding global predictive performance, some models require orders of magnitude more samples than others to reach a certain likelihood or RMSE. The BNN seems most capable as a global metamodel due to its low RMSEs and heterogeneous uncertainties resulting in the highest likelihoods. However, using evenly distributed samples, none of the models performed well on the metrics used to evaluate the performance for typical tasks in scenario-based testing, such as boundary identification.

To build models that perform well for these tasks, we introduced a scalable, intuitive approach to select samples. As a result, the task-specific performance can be greatly improved to a degree where it seems evident that the appropriate selection of samples is more important than the choice of metamodels. In general, the predictive performance seems to depend more on a system's complexity than the number of inputs it is modeled with. This implies that it is not necessary to rigorously limit the number of inputs during the modeling stage, which reduces the risk of omitting possibly crucial influences of the environment on an examined system.

While the acquisition function strongly influences the final model quality, for many applications, choosing an appropriate metamodel is still crucial. From a macroscopic view, the BNN seems to be the best metamodel indicated by high likelihoods and low RMSEs. However, there often is a small fraction of samples where the BNN's predictions are rather poor. The less flexible models are more conservative, i.e., they are not especially good for \textit{most} test cases but achieve reliable performance for \textit{almost all} test cases. Among these models, the GP seems to be a very good choice due to its widespread use and solid performance for all performance indicators.

Overall, metamodels are well-suited for prioritizing test cases. However, metamodels' predicted risks should be treated with caution. The choice of metamodels and exploration strategies always entails a tradeoff between efficiency and reliability. This tradeoff needs to be discussed not only methodologically but also ethically to provide society with a deeper understanding of AV safety validation and its limits.

We conclude that in early development stages, where data is easily available and which may be quite explorative because the system and the operational design domain may not be fully defined, flexible models like BNNs can significantly contribute to characterizing AVs as efficiently and completely as possible. In advanced stages, where less data is available and model calibration is more important, more reliable models like GPs can create evidence for the validation and release of AVs.

\section{Future Work}\label{sec:future_work}
There are many directions for future work. First, considering prior assumptions or the likelihood of concrete scenarios could speed up our iterative approach to some extent. However, it also introduces the risk of introducing a bias towards the experience of human drivers and overseeing scenarios that are challenging for AVs. For the task of boundary identification, considering the uncertainty of the predictions in the acquisition function (e.g., as done in \cite{batsch_scenario_2021}) could help to keep the focus of samples that lie exactly on the safety boundary and shift it towards samples with uncertain outcomes.

Furthermore, common safety metrics are created for human intuition and not for the purpose of building metamodels. Metrics created with a focus on properties such as continuity would allow modeling a system's behavior more easily, especially for less flexible metamodels. Here, the likelihoods used for predictions play a role as well. For example, Gaussian likelihoods are not well suited to predict metrics that converge to zero (e.g., the $\mathrm{TTC_{min}}$). Other likelihoods, possibly with wider tails, could improve the predictions and calibration.

Most importantly, we want to apply the developed concepts to more realistic test setups with reduced data availability. Across multiple test setups, metamodels can support the selection of test cases, while the results are used to continuously refine the metamodels. This direction has been explored in \cite{huang_synthesis_2018}, assuming a common input space of the test setups. In~\cite{winkelmann_transfer_2022}, we consider changing inputs, e.g., weather conditions in proving ground tests could correspond to noise in a 2D simulation. This direction offers great potential to capture and utilize available knowledge and test setups more efficiently.

\begin{table}[!ht]
	\renewcommand{\arraystretch}{1.1}
	\begin{center}
		\caption{Metamodel Hyperparameters}
		\label{tab:hyperparameters}
		\begin{tabular}{|l|l|l|}
			\hline
			\textbf{Model}                    & \textbf{Parameter}                    & \textbf{Value}                                           \\

			\hline 
			\multirow{2}{*}{$\mathrm{GP}$}    & Implementation                        & GPyTorch~\cite{gardner_gpytorch_2018}                    \\
			\cline{2-3}
			                                  & Kernel                                & $\mathrm{Matern}_{\nu=0.5}$, $l$ per input               \\
			\cline{2-3}
			                                  & Learning Rate                         & $0.1$                                                    \\
			                                  & Early Stopping                        & 200 epochs with $\Delta = 0.01$                          \\
			\cline{2-3}
			                                  & $t_\mathrm{train}$                    & \scriptsize{5.8 s ($N_S = 500$), 129 s ($N_S = 10^4$)}   \\
			                                  & $t_\mathrm{sampling}/\mathrm{sample}$ & \scriptsize{80 ms (Lanekeeping), 36 ms (Parking)}        \\

			\hline 
			\multirow{2}{*}{$\mathrm{DGP_2}$} & Implementation                        & GPyTorch~\cite{gardner_gpytorch_2018}                    \\
			\cline{2-3}
			                                  & Kernel                                & Rational Quadratic, $l$ per input                        \\
			                                  & Shape                                 & [$N_I$, $N_O$]                                           \\
			                                  & Inducing Points                       & 100                                                      \\
			                                  & Mixture Components                    & 9                                                        \\
			\cline{2-3}
			                                  & Learning Rate                         & $0.01$                                                   \\
			                                  & Early Stopping                        & 200 epochs with $\Delta = 0.01$                          \\
			\cline{2-3}
			                                  & MC Samples                            & 50                                                       \\
			\cline{2-3}
			                                  & $t_\mathrm{train}$                    & \scriptsize{71 s ($N_S = 500$), 1259 s ($N_S = 10^4$)}   \\
			                                  & $t_\mathrm{sampling}/\mathrm{sample}$ & \scriptsize{162 ms (Lanekeeping), 72 ms (Parking)}       \\

			\hline 
			\multirow{2}{*}{$\mathrm{ET}$}    & Implementation                        & scikit-learn 0.24.0~\cite{pedregosa_scikit-learn_2011}   \\
			\cline{2-3}
			                                  & Estimators                            & $100$                                                    \\
			                                  & Min Samples Split                     & 4                                                        \\
			                                  & Max Features                          & $N_I$                                                    \\
			\cline{2-3}
			                                  & $t_\mathrm{train}$                    & \scriptsize{0.27 s ($N_S = 500$), 0.54 s ($N_S = 10^4$)} \\
			                                  & $t_\mathrm{sampling}/\mathrm{sample}$ & \scriptsize{29 ms (Lanekeeping), 26 ms (Parking)}        \\

			\hline 
			\multirow{2}{*}{$\mathrm{BNN}$}   & Implementation                        & Concrete Droput~\cite{gal_concrete_2017}                 \\
			\cline{2-3}
			                                  & Shape                                 & [$N_I$, 180, 110, 430, 830, 160, $2 N_O$]                \\
			                                  & Activation                            & Hardswish                                                \\
			                                  & Concrete Dropout                      & $\tau = 7$, $l = 0.0005$                                 \\
			\cline{2-3}
			                                  & Minibatch Size                        & 8                                                        \\
			                                  & Learning Rate                         & $0.001$ ($* 0.2$ each 10k minibatches)                   \\
			                                  & Early Stopping                        & 20k minibatches with $\Delta = 0.1$                      \\
			\cline{2-3}
			                                  & MC Samples                            & 50                                                       \\
			\cline{2-3}
			                                  & $t_\mathrm{train}$                    & \scriptsize{1699 s ($N_S = 500$), 1765 s ($N_S = 10^4$)} \\
			                                  & $t_\mathrm{sampling}/\mathrm{sample}$ & \scriptsize{43 ms (Lanekeeping), 54 ms (Parking)}        \\
			\hline
		\end{tabular}
	\end{center}
\end{table}

\appendix[Driving Scenario Input Output Dataset]\label{sec:dataset}
Tabular results, descriptions, and videos of the investigated scenarios are available at \url{https://github.com/wnklmx/DSIOD}. We invite researchers to add similar datasets to this repository.

\section*{Acknowledgment}
The authors would like to thank Mike Hartrumpf, Constantin Vasconi, David Seidel, Roland Kallweit, and Matthias Butenuth of IAV GmbH for their feedback and support.

\ifCLASSOPTIONcaptionsoff
	\newpage
\fi



\bibliographystyle{IEEEtran}
\bibliography{IEEEabrv,./refs/refs}

%

\begin{IEEEbiography}[{\includegraphics[width=1in,height=1.25in,clip,keepaspectratio]{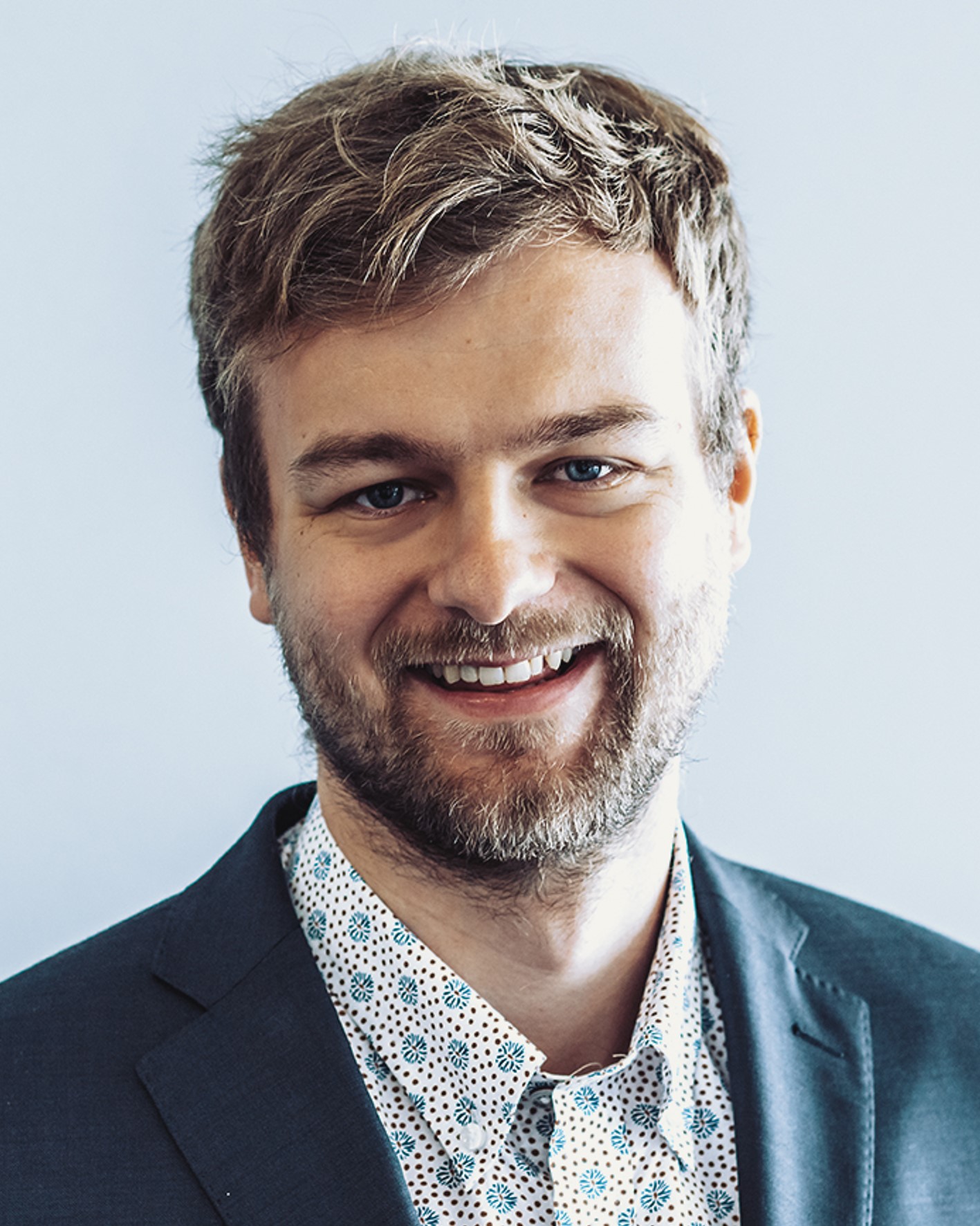}}]{Max Winkelmann}
	is currently a doctoral student at IAV GmbH, Berlin, Germany and pursuing the Dr.-Ing. degree in automotive engineering at Technische Universität Berlin, Berlin, Germany. He received the M.Sc. degree in electrical engineering in 2020 from Technische Universität Berlin. With roots in electrical engineering, control, and automation, his interests steadily evolved towards the fields of robotics, computer vision, and machine learning. Supported by this background, his motivation is to gain a global understanding of complex technical systems. His current research is concerned with the question of how an expert's understanding of a system can be complemented by machine learning to finally unite qualitative and quantitative insights.
\end{IEEEbiography}

\begin{IEEEbiography}[{\includegraphics[width=1in,height=1.25in,clip,keepaspectratio]{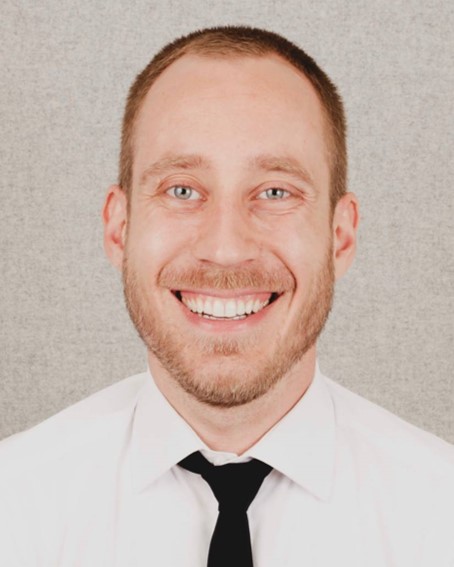}}]{Mike Kohlhoff}
	received the Ph.D. degree from the University of Oxford, Oxford, United Kingdom in 2017. A physicist by training, he has worked on automated inspection of transport systems, quantum sensors, and sustainable hydrogen production. At IAV GmbH, his focus was on the development of virtual test solutions for automated driving systems. Currently, he works on machine learning problems and advanced modeling for virtual representations of physical systems. Further research interests include concept work for mobility solutions in urban environments and NISQ quantum computation algorithms.
\end{IEEEbiography}

\begin{IEEEbiography}[{\includegraphics[width=1in,height=1.25in,clip,keepaspectratio]{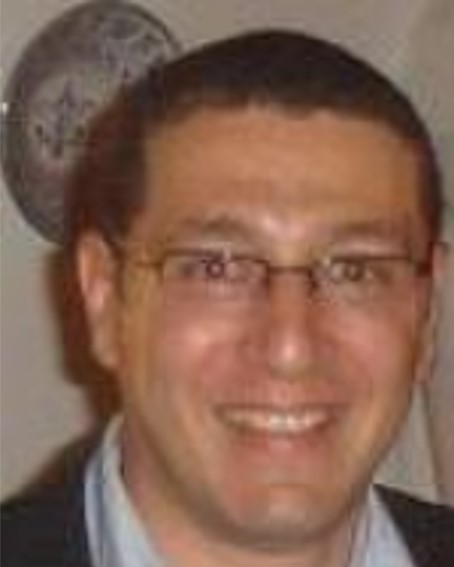}}]{Hadj Hamma Tadjine} (Senior Member, IEEE) received the Dr. rer. nat. degree in computer science from the Clausthal University of Technology, Zellerfeld, Germany. From 2000 to 2004, he was assistant professor at the Clausthal University of Technology. From 2004 to 2006, he was assistant professor at CUTEC Institute GmbH, Zellerfeld, Germany. From 2006 to 2008, he was responsible for advanced driver assistance systems at Hella Aglaia, Berlin, Germany. From 2008 to 2010, he was responsible for advanced driver assistance systems and parking assistance systems at IAV GmbH, Berlin, Germany. Currently, he is the businesses director for technical strategy in the area of intelligent driving systems. He has a track record of fundamental research on these topics documented by numerous publications by IEEE, VDI, and SAE. He is the editor and editor in chief of different international journals.
\end{IEEEbiography}

\begin{IEEEbiography}[{\includegraphics[width=1in,height=1.25in,clip,keepaspectratio]{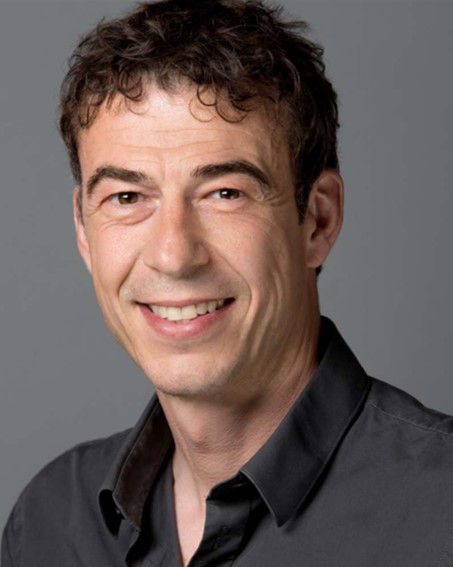}}]{Steffen Müller}
	is university professor and ``Einstein Professor'' at Technische Universität Berlin, Berlin, Germany. He is the head of the Chair of Automotive Engineering of the Faculty of Mechanical Engineering and Transport Systems. He received the Dipl.-Ing. degree in astronautics and aerospace engineering in 1993 and the Dr.-Ing. degree in 1998 from Technische Universität Berlin. From 1998 to 2000, he was a project manager at the ABB Corporate Research Center, Heidelberg, Germany. He finished the postdoctoral research at University of California, Berkeley in 2001. From 2001 to 2008, he had taken up different leading positions at the BMW Research and Innovation Centre. From 2008 to 2013, he was founding and leading the Chair for Mechatronics in Engineering and Vehicle Technology at Technical University of Kaiserslautern, Kaiserslautern, Germany.
\end{IEEEbiography}




\enlargethispage{0.29in}

\end{document}